\documentclass[runningheads]{llncs}
\usepackage[T1]{fontenc}
\usepackage{graphicx}
\usepackage{subcaption}

\newcommand{\rulesep}{\unskip\ \vrule\ }

\begin{document}

\title{Topics in Contextualised Attention Embeddings}
\author{%
Mozhgan Talebpour\inst{1},
Alba Garcia Seco de Herrera\inst{1},
Shoaib Jameel\inst{2}
}%
\institute{
Computer Science and Electronic Engineering, University of Essex, United Kingdom.
\and
Electronics and Computer Science, University of Southampton, United Kingdom.
\email{\{mozhgan.talebpour, alba.garcia\}@essex.ac.uk, M.S.Jameel@southampton.ac.uk}
}

\maketitle

\begin{abstract}
Contextualised word vectors obtained via pre-trained language models encode a variety of knowledge that has already been exploited in applications. Complementary to these language models are probabilistic topic models that learn thematic patterns from the text. Recent work has demonstrated that conducting clustering on the word-level contextual representations from a language model emulates word clusters that are discovered in latent topics of words from Latent Dirichlet Allocation. The important question is how such topical word clusters are automatically formed, through clustering, in the language model when it has not been explicitly designed to model latent topics. To address this question, we design different probe experiments. Using BERT and DistilBERT, we find that the attention framework plays a key role in modelling such word topic clusters. We strongly believe that our work paves way for further research into the relationships between probabilistic topic models and pre-trained language models.
\end{abstract}

\section{Introduction}
%
Pre-trained language models (PLMs), e.g., ELMo~\cite{DBLP:journals/corr/abs-1802-05365}, Generative Pre-trained Transformer (GPT)~\cite{radford2018improving}, PaLM~\cite{chowdhery2022palm}, and Bidirectional Encoder Representations from Transformers (BERT)~\cite{devlin2018bert} are pre-trained using large amounts of text data~\cite{lai2020context}, for instance, BERT has been pre-trained on the BookCorpus and Wikipedia collections. During the domain-independent pre-training process, these models encode a variety of latent information, for instance, semantic and syntactic properties~\cite{warstadt2019investigating}, as a result, these models can make reliable predictions even under a zero-shot setting in different applications~\cite{heo2022mcbert,ristoski2021kg,rogers2020primer}. While the pre-training process is computationally~\cite{turc2019well} and financially expensive~\cite{strubell2019energy}, these models can be cheaply fine-tuned to reliably handle different downstream tasks such as document classification~\cite{adhikari2019docbert} and information retrieval~\cite{yang2019simple,trabelsi2021neural}, a process that is commonly referred to as transfer learning~\cite{mozafari2019bert}. For instance, BERT has shown strong performance in natural language understanding~\cite{zhang2020semantics}, text summarisation~\cite{lamsiyah2021unsupervised}, document classification~\cite{chalkidis2019large} and other Natural Language Processing (NLP) downstream applications~\cite{rogers2020primer}.

Another class of models that continues to dominate the text mining landscape are probabilistic topic models (PTMs)~\cite{blei2003latent,blei2012probabilistic}. These models are probabilistic approaches toward determining dominant topics in a text corpus in a completely unsupervised way. A latent topic is described as a probability distribution of words. Latent Dirichlet Allocation (LDA)~\cite{blei2003latent} is a popular model for discovering topics. In LDA, the model learns to represent a document as a mixture of latent topics and each topic is represented by a mixture of words. When LDA is viewed as a matrix factorisation model, given a term document co-occurrence matrix as input and the number of topics, the model factorises the matrix into two low-dimensional matrices that are word topic and document topic representations. The word topic matrix captures the importance of words in the vocabulary of each topic whereas the document topic matrix captures the topic distribution in every document. While LDA has been a popular model that is based on Bayesian learning, a class of linear algebra-based model called Non-negative Matrix Factorisation (NMF)~\cite{wang2012nonnegative,lee1999learning} has become equally popular to learn topics~\cite{de2016equivalence}.

In~\cite{rogers2020primer}, the authors dissected BERT to understand the property of every layer. They find that lower layers, i.e., layer 1 or 2 capture the linear word order, while the BERT's middle layers learn the syntactic information reliably and the higher layers capture the contextualised information. The authors in~\cite{thompson2020topic} and~\cite{sia2020tired} showed that BERT word embedding clustering via simple algorithms such as k-means results in word clusters as if they are learned by a topic model. The authors conducted a series of qualitative probe experiments to find out that most of the word clusters of BERT resemble what is often discovered by the LDA model. While these studies make relevant observations, what is not well studied is how the topic information is encoded at the time of pre-training given that BERT or any other contextual language model is not designed to model topical word clusters. In this work, by conducting different probe experiments, we answer how BERT and DistilBERT~\cite{sanh2019distilbert} can capture clusters of words that resemble what is learnt by topic models. We find that it is the attention \cite{bibal-etal-2022-attention,brunner2019identifiability} mechanism in these language models that plays a key role in modelling what resembles word topics as discovered by the topic model.

\section{Related work}\label{previous work}
%
The main goal of PLMs~\cite{min2021recent} is to simulate human language understanding by finding the most probable words sequence and patterns. The traditional language model used probability distribution to predict the next word, but they were not very scalable such as those based on unigram, bigram or trigram language models~\cite{ponte2017language}. The recently developed PLMs are trained using large amounts of text data where some of them exploit a strategy called masked language modelling in a self-supervised way. Once these models have been trained, they have been applied in a wide variety of applications. The key advantage of PLMs is that they can be applied on different downstream tasks~\cite{edunov2019pre} reliably.

\begin{figure}[t]
\begin{subfigure}[t]{0.50\textwidth}
    \includegraphics[width=\textwidth]{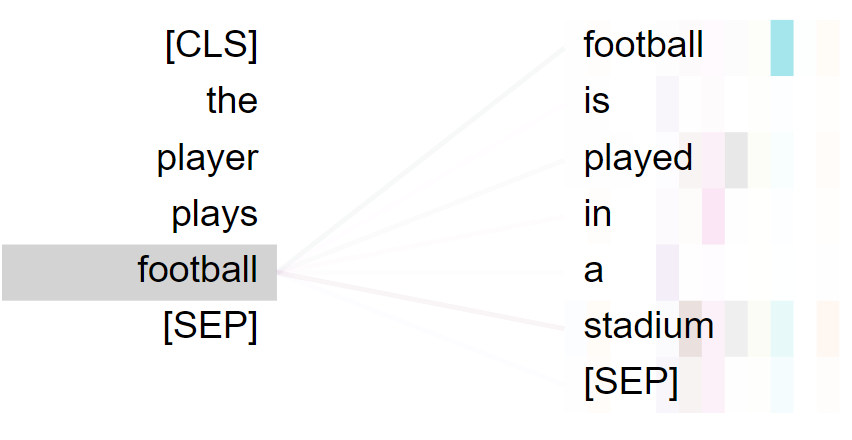}
    \caption{Attention mechanism in BERT via visualisation in Layer 12. We observe that words that are central to the context are assigned high attention weights.}
    \label{fig:visualise_attention_left}
\end{subfigure}
\rulesep
\begin{subfigure}[t]{0.50\textwidth}
    \includegraphics[width=\textwidth]{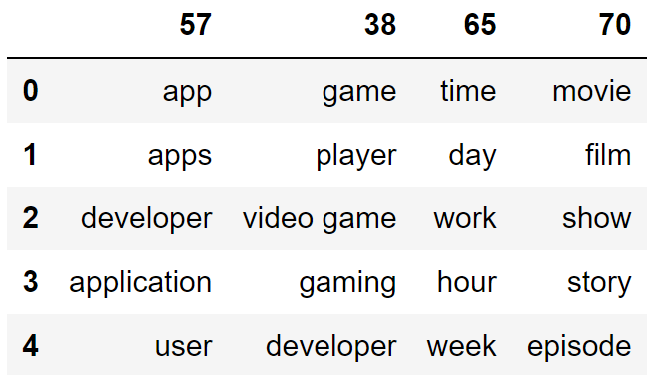}
    \caption{Words, ordered by decreasing probability, obtained from the LDA model.}
    \label{fig:visualise_attention_right}
\end{subfigure}
\end{figure}

BERT has been developed with stacked transformers~\cite{vaswani2017attention} layers where each layer captures different properties in text data, e.g., some layers are ideal to capture semantic information \cite{voita2019bottom,tenney2019bert}. Transformers consist of encoder-decoder structures. The encoder transforms the sequence of input tokens into a high-level dimension. Decoder predicts input data from encoder~\cite{futami2020distilling}. However, in BERT only the encoder part of transformers has been used. There is an important concept in BERT called attention that assigns weights to different input features given their importance in the underlying task. One example is: given the text about \textit{cats}, the model will pay more attention, via attention weights, to words such as \textit{fur}, \textit{eyes}, etc. BERT's attention has also been studied in~\cite{clark2019does} where the authors find that different attention heads focus on different aspects of language, e.g., they find that heads direct objects of verbs, determiners of nouns, objects of prepositions, and objects of possessive pronouns with far greater accuracy. While they have studied the syntactic and semantic information encoded in different attention heads, they have not separately probed latent topics as learned by the topic models such as LDA and NMF. In Figure~\ref{fig:visualise_attention_left}, we depict how attention works obtained via a popular visualisation tool\footnote{https://github.com/jessevig/bertviz}. We input two sentences in sequence, where the first sentence ``The player plays football.'' is followed by the second sentence ``Football is played in a stadium.'', and both describe the sport \textit{football}. The visualisation tool depicts the case when we select the token ``football'' in the first sentence and how other semantically related tokens such as ``football'', ``stadium'', and ``played'' are highlighted with high attention weights.
%

Topic modelling is a machine learning technique that automatically discovers hidden topics in unlabelled data. A topic is defined as a probability distribution of words. While these topic models have been inspired by the latent concept-based models such as Latent Semantic Analysis (LSA)~\cite{deerwester1990indexing} and Probabilistic Latent Semantic Analysis (pLSA)~\cite{hofmann2013probabilistic}, Latent Dirichlet Allocation (LDA)~\cite{blei2003latent} has been widely applied to discover latent topics because it addressed some of the fundamental challenges in LSA and pLSA such as scaling on large datasets and overfitting. While in~\cite{levy2014neural}, the authors demonstrated that static word embeddings are related to SVD, which is the core algorithm used in LSA, what we demonstrate here is that models such as PLMs implicitly learn latent topic information as encoded by the PTMs.

LDA has been trained considering the exchangeability~\cite{foti2013survey} assumption meaning that word order does not matter in a document. These models describe documents as a mixture of words and each document comprises a mixture of topics defined by the user. Note that BERT does not model document-level information; there are extensions such as Sentence-BERT (SBERT)~\cite{reimers2019sentence} to model documents.

In Figure~\ref{fig:visualise_attention_right}, we depict a typical output obtained from LDA using a freely available online topic modelling visualisation tool\footnote{https://pyldavis.readthedocs.io/en/latest/index.html}. We can observe from this output that there are five top-ranked probability words in some topics that are indexed by topic labels as discrete numbers. From topic index 57, we can infer that the topic describes computer or mobile applications and their development. Topic number 38 describes video gaming.

BERT has demonstrated state-of-the-art results in many NLP downstream tasks, such as natural language inference and information retrieval. Some previous studies have emphasised the importance of contextual information as an additional feature of topic modelling. In~\cite{bianchi2020pre}, for example, the importance of sentence contextual representation and neural topic model was investigated. In SBERT~\cite{reimers2019sentence}, embedding representation was used as the input to the prodLDA~\cite{srivastava2017autoencoding} neural topic model. If an input document length exceeded the SBERT predefined length, the rest of the document would be omitted. Despite this limitation, the model produced a higher coherence score when compared to Bag-of-Words (BoW) representation embedding. Some other studies have focused on how, and if, adding topic modelling information to a BERT model can lead to an improvement in its performance. In a study conducted by Peinelt et al.,~\cite{peinelt2020tbert}, they have used topic modelling to improve the BERT performance of semantic similarity domain applications like question answering. They have used BERT-base final layer's \textit{[CLS]} token embedding as the corresponding embedding of an input document. Wang et al.,~\cite{wang2021deriving} have argued that BERT contextual embedding can be improved by adding topical information to it. In their study, BERT embedding was derived from topics in the corpus. The findings of this research suggest that a word vector representation is equal to the weighted average of different topical vectors. If a topic has high importance in a corpus, words that are related to that topic gain higher importance.

In another related research conducted by~\cite{iida2004lecture}, topical text classification was applied to a scientific domain dataset. The authors compared the findings of their research with SciBERT~\cite{beltagy2019scibert}, which is a pre-trained language model based on BERT, but on scientific documents. Concatenation of BERT embedding and document topic vector was used as an input to a two-layer feed-forward neural network. In a recent study,~\cite{thompson2020topic}, the role of BERT embedding was examined from a different perspective. This research argued that clustering token-level BERT embedding shares many similarities with topic modelling. The authors used different PLMs such as BERT, GPT-2~\cite{radford2019language} and RoBERTa's~\cite{liu2019roberta} last three layers of embedding. This work found that except RoBERTa, BERT and GPT-2 word-level clustering resulted in clusters that resemble close to those obtained using the LDA model. While LDA learns topics as a probability distribution of words, the word clusters obtained by clustering token-level embeddings in PLM cannot be confused with a probability distribution of words. What the authors showed is that there are some similarities between the word clusters of a PTM when compared with the clusters obtained from a PLM.

While the works mentioned above demonstrate important relationships between PTMs and PLMs, what is currently lacking is a further understanding of how latent topics are encoded in the PLM vectors and what component helps in encoding this information. There are works mentioned above that have trained latent topics with pre-trained language models in a unified way. The question is whether it is needed to learn latent topics with pre-trained language models again. While these works have shown quantitative improvements, it is unclear how latent topics are helping them improve upon the results.

\section{Probe Tasks}\label{methodology}
%
The problem that we intend to study in this paper is whether latent topic information is automatically encoded in contextualised word embeddings. While it is not explicitly evident that latent topic information is encoded, we must design probe tasks. Our key goal is thus to understand how PLMs such as BERT and DistilBERT can discover word clusters that are often discovered by PTMs when they are not specifically designed to model such information. To this end, we first chose to study in more detail the role that attention heads play in the PLM model. It is because just as in a topic model, words that are central to the document's global context are assigned a high probability and words that are central to a topic are assigned a high probability. For instance, if the document is about ``sports'', words such as ``football'', ``goal'', and ``player'' will have a high probability in that document. Similarly, these words will occur with a high probability in the topic that is about sports. The attention mechanism too shows similar behaviour in the document where words that are central within the given contextual window are assigned high attention weights. Attention weight specifies the importance of a particular word when it is accompanied by other words~\cite{clark2019does} in a certain pre-defined contextual window.

We consider BERT-base uncased and DistilBERT-based uncased models as our PLMs because of their popularity and computational ease. We also know that the LDA model outputs word and document topic representations~\cite{blei2003latent}. Given the number of factors or latent dimensions, NMF factorises the co-occurrence matrix into two low-dimensional matrices where one matrix encodes word clusters and the other matrix encodes document clusters. Since language models capture word-level patterns, we thus choose word topics in LDA and NMF. Since both LDA and NMF can explicitly be assigned to soft clusters based on their probability values, in the case of the attention representations, we must cluster them using a soft clustering algorithm. This would help us produce word clusters with cluster assignments.

There are other components that we could also study such as the role played when different transformers layers when stacked together. However, previous studies have already found out that the different layers capture different properties of text data, e.g., in BERT, lower layers capture linear word order, middle layers capture syntactic information whereas higher layers capture semantic information. None of these studies has found that word clusters resembling latent topics are also modelled by one of these layers after thorough experimentation. As a result, given their findings, we focus on the attention heads in PLMs first.

In BERT-base, there are 12 layers, each layer containing 12 attention heads. The attention head computes attention weights between all pairs of word combinations in an input sentence. Attention weight can be interpreted as an important criterion when considering two words simultaneously. For example (weather, sunny) pair's attention weight is higher than the (weather, desk) pair. It is because when BERT is trained on billions of tokens (weather, sunny) combinations occurred more frequently than other words such as ``desk''. Similarly, in the LDA model, if words such as ``sports'' and ``football'' occur, they will be assigned a high probability value in the word topic. DistilBERT is also based on the BERT-base model but is much lighter weight with respect to its parameters. It has been obtained after a process known as knowledge distillation~\cite{lopes2017data,hahn2019self} where the original bigger model known as the teacher was used to train the lighter-weight compressed student model to mimic its behaviour. It was found that in the case of DistilBERT, it retained most of BERT's advantages with a much-reduced parameter set.

Using two publicly available benchmark datasets, we conduct two different probe tasks to demonstrate the generalisability of our findings. In the first probe task, we conduct word-level clustering on the representations obtained from PTM and PLM models and compute the coherence measure which has been popularly used in topic models to evaluate the quality of the topics. In the case of the language models, we extract attention weights from each layer of the model and we obtain the word-attention representations for every word in the vocabulary. We then cluster these attention vectors using a clustering algorithm where the attention vectors are used as features. Through this attention clustering, we expect that words that are semantically related are clustered in one cluster. The motivation is that if the word clusters contain thematically related words, the clusters will demonstrate a high coherence measure. While there have been debates around the usefulness of coherence measure \cite{hoyle2021automated}, in our study, we use the same measure to compare all models quantitatively.

We intend to probe if there is a comparable coherence performance between a layer of PLM and the word-topic representations obtained from PTM. By comparable, we mean whether the coherence results are numerically close to each other. If the coherence results are comparable, we can expect that in terms of the thematic modelling of words, the language model and the topic models are learning semantically related content. While the coherence probe task might not completely be relied upon, we design an additional probe task to find out the word overlaps between the word clusters obtained from the PLMs and PTMs. Our motivation is that if the coherence value between the clusters is high then there must be a reliable overlap between the words in the clusters. Since the higher layers, 10, 11 and 12 in the case of the BERT-base model capture semantic information more than the lower layers, we expect that the clusters of words in the high layers of the language model will show higher commonality with those clusters that are learnt by the PLMs.
%

\subsection{Experimental Settings}
%
\emph{Datasets}: We have used 20 NewsGroups (20NG) and IMDB datasets which are two popular datasets commonly used in the text mining community. The 20NG dataset contains about 18,000 documents in 20 news categories after removing duplicate and empty instances. IMDB dataset contains 50,000 movie reviews that have been labelled as positive or negative. The 20NG dataset contains several long documents whereas IMDB contains relatively short documents with relatively more noisy text.

\emph{Text preprocessing}: In the case of the PTMs, we have followed a common pre-processing strategy such as the removal of the stop words, the removal of punctuation, and non-ASCII characters. Through our experiments, we have found that if we do not remove stopwords from text, they tend to dominate most of the topics including increasing the dimensionality of the semantic space resulting in high space and time complexities. While some workaround have been proposed to model natural language using PTMs such as using asymmetric priors, they can be computationally intensive on large datasets \cite{wallach2009rethinking}. In the case of the PLMs, we let the default pre-processor handle pre-processing, for instance, the BERT-base model has the WordPiece tokenizer. Using NLTK~\cite{bird2009natural}, we conducted sentence segmentation.

\emph{PLM attention weights}: For every word in the vocabulary, we obtain the word attention weights from the BERT-base uncased and DistilBERT models. As BERT uses wordPiece tokenisation, if tokenised sentence length is more than 512 tokens, the input sentence is split which is common in the literature. Attention weights of all tokens in a sentence would be stored. If a word appears in different sentences, the average of all words' attention is used which is also commonly done including taking their average embedding of their word pieces~\cite{wu2016google}. We have obtained attention weights from every layer of BERT. BERT attention weight has been defined as an average of all attention heads in each layer.

We have obtained attention weights from the vanilla BERT-base model. Besides that, we have also obtained attention weights from the fine-tuned version of the BERT model to gauge the role fine-tuning might play in the process. Fine-tuning was done on the text classification task using labels associated with labelled instances in the 20NG and IMDB datasets. Through cross-validation in the fine-tuning process, we present the results of the best-performing model on the test set with the ideal model parameters obtained via a 30\% held-out dataset. We have followed the same configuration with the vanilla DistillBERT-base model.

\emph{Topic modelling}: We have used the Latent Dirichlet Allocation (LDA) model implemented in Gensim~\cite{rehurek2011gensim} to discover latent topics in our datasets. In the 20NG dataset, we have varied the number of topics from 20 to 200. In the IMDB dataset, we varied the number of topics from 2 to 30 which gave us better results. We have used the NMF model implemented in Gensim. According to \cite{wei2006lda}, larger datasets tend to have more topics than smaller ones. As a result, we have chosen different topic pools in different datasets. We have not chosen the number of topics to be equal to the dimensionality of the word vectors obtained from PTMs because PTMs tend to encode a variety of information in their vectors, e.g., syntactic and semantic information. Besides that, having many topics larger than what we have chosen above tends to result in sub-optimal latent topics leading to the deterioration of performance.

\emph{Clustering}: We have used the soft Gaussian Mixture Models (GMM)~\cite{bishop2006pattern} clustering algorithm on the embeddings obtained from PLMs. LDA is already a soft clustering model where probability values are used to assign soft clusters to instances~\cite{xu2003document}. In LDA, we can automatically obtain the word-topic assignments based on the probability values of words in each topic which is also true for clusters obtained via the GMM model. We used GMM because its implementation is widely and freely available in different software libraries.

\emph{Evaluation}: In topic modelling, coherence measure has been widely used to evaluate the quality of the latent topics~\cite{mimno2011optimizing}. Coherence score ``c-v'' has been used in our setting which is available in the Gensim library. This measure has been adapted from the work of Roder et. al.~\cite{roder2015exploring}. We use coherence to measure the semantic relatedness of tokens in the word clusters obtained from both PLM and PTM models. We also use the number of word overlaps between the top-k words in clusters obtained from the two models to gauge the word overlaps among the clusters. We set \(k=20\) which gives a reliable trade-off between selecting the most thematically related top-k words and not choosing (general or noisy) words with low probability estimated in the word clusters. To compute the word overlap values, for every topic in PTM and every layer's word cluster in PLMs, we computed the overlap between the top-k words, followed by computing the ``mode'' value. While there are metrics such as entropy and exclusivity \cite{thompson2020topic}, we will use these metrics in the extended version of this paper.
\begin{table}[t]
\centering
\scalebox{0.7}{
\begin{tabular}{llll}
\hline
\hline
\multicolumn{4}{c}{\textbf{20 Newsgroups}}                              \\ \hline
\hline
\multicolumn{2}{c}{\textbf{LDA}}     & \multicolumn{2}{c}{\textbf{NMF}} \\ \hline
\# topics & \multicolumn{1}{l|}{c-v} & \# topics          & c-v         \\ \hline
20        & \multicolumn{1}{l|}{\textbf{0.518}}    & 20                 &     0.478        \\
30        & \multicolumn{1}{l|}{0.487}    & 30                 &    \textbf{0.504}     \\
50        & \multicolumn{1}{l|}{0.504}    & 50                 &    0.484         \\
100       & \multicolumn{1}{l|}{0.474}    & 100                &    0.453         \\
150       & \multicolumn{1}{l|}{0.470}    & 150                &    0.455         \\
200       & \multicolumn{1}{l|}{0.473}    & 200                &        0.474     \\ \hline
\end{tabular} %
}
\quad
\scalebox{0.7}{
\begin{tabular}{llll}
\hline
\hline
\multicolumn{4}{c}{\textbf{IMDB}}                              \\ \hline
\hline
\multicolumn{2}{c}{\textbf{LDA}}     & \multicolumn{2}{c}{\textbf{NMF}}   \\ \hline
\# topics & \multicolumn{1}{l|}{c-v} & \# topics          & c-v         \\ \hline
2        & \multicolumn{1}{l|}{0.363}    & 2                 &     0.276        \\
5        & \multicolumn{1}{l|}{0.364}    & 5                 &      0.275       \\
10        & \multicolumn{1}{l|}{0.370}    & 10                 &    0.299         \\
20       & \multicolumn{1}{l|}{\textbf{0.461}}    & 20                &      \textbf{0.300}       \\
30       & \multicolumn{1}{l|}{0.437}    & 30                &      0.299       \\
\hline
\end{tabular}
}
\caption{Coherence results for LDA and NMF models.}
\label{20ng_coherence}
\end{table}
%
\begin{table*}[t]
\centering 
\scalebox{0.6}{
\begin{tabular}{c c c c c c c c c c c} 
\hline\hline 
Layer & VB30 & VB50 & VB100 & VB150 & VB200 & FT30 & FT50 & FT100 & FT150 & FT200 \\ [0.5ex] 
\hline 
1 &0.360 & 0.502 & 0.489 & 0.481 & 0.343 & 0.333 & 0.477 & 0.502 & 0.466 & 0.330 \\ 
2 &0.346 & 0.480 & 0.463 & 0.464 & 0.334 & 0.327 & 0.479 & 0.480 & 0.462 & 0.333 \\
3 & 0.329 & 0.450 & 0.453 & 0.448 & 0.323 & 0.315 & 0.466 & 0.450 & 0.459 & 0.324 \\
4 &0.328 & 0.466 & 0.461 & 0.452 & 0.332 & 0.324 & 0.466 & 0.466 & 0.461 & 0.332\\
5 & 0.33 & 0.460 & 0.448 & 0.449 & 0.324 & 0.325 & 0.458 & 0.460 & 0.453 & 0.329\\
6 & 0.33 & 0.459 & 0.455 & 0.451 & 0.318 & 0.347 & 0.466 & 0.459 & 0.460 & 0.337\\
7 & 0.337 & 0.478 & 0.471 & 0.454 & 0.325 & 0.346 & 0.495 & 0.478 & 0.479 & 0.347\\
8 & 0.347 & 0.469 & 0.468 & 0.470 & 0.336 & 0.353 & \textbf{0.508} & 0.469 & \textbf{0.496} & 0.359\\
9 & 0.346 & 0.486 & 0.474 & 0.471 & 0.344 & \textbf{0.370} & \textbf{0.508} & 0.486 & 0.494 & 0.360\\
10 & 0.373 & 0.480 & 0.483 & 0.476 & 0.360 & 0.368 & 0.502 & 0.480 & 0.494 & 0.358 \\
11 & 0.369 & \textbf{0.503} & 0.489 & 0.481 & 0.360 & 0.357 & 0.483 & \textbf{0.503} & 0.489 & \textbf{0.361}\\
12 & \textbf{0.373} & 0.502 & \textbf{0.489} & \textbf{0.484} & \textbf{0.363} & 0.364 & 0.485 & 0.502 & 0.480 & 0.355 \\ [1ex] 
\hline 
\end{tabular} %
\begin{tabular}{c c c c c c c c c c c} 
\hline\hline 
Layer & VB2 & VB5 & VB10 & VB20 & VB30 & FT2 & FT5 & FT10 & FT20 & FT30\\ [0.5ex] 
\hline 
1 & 0.411 & 0.390 & 0.365 & 0.358 & 0.355 & 0.455 & 0.442 & 0.372 & 0.348 &0.333 \\
2 & 0.473 & 0.447 & 0.374 & 0.347 & 0.352 & 0.469 & \textbf{0.459} & 0.420 & 0.391 & 0.384 \\
3 & 0.480 & 0.414 & 0.418 & 0.385 & 0.366 & 0.501 & 0.452 & 0.415 & \textbf{0.412} & \textbf{0.404} \\
4 & \textbf{0.586} & 0.478 & 0.444 & 0.421 & 0.404 & 0.457 & 0.388 & 0.386 & 0.383 & 0.380 \\
5 & 0.583 & \textbf{0.490} & 0.439 & 0.426 &  0.422 & 0.410 & 0.383 & 0.350 & 0.359 & 0.356 \\
6 & 0.563 & 0.477 & \textbf{0.471} & 0.429 &0.405 & 0.452 & 0.438 & 0.396 & 0.374 & 0.357\\
7 & 0.546 & 0.485 & 0.431 & 0.425 & 0.416 & 0.489 & 0.403 & 0.399 & 0.374 & 0.366 \\
8 & 0.510 & 0.438 & 0.428 & 0.414 & 0.415 & \textbf{0.514} & 0.427 & 0.395 & 0.397 & 0.369 \\
9 & 0.452 & 0.410 & 0.393 & 0.383 & 0.373 & 0.438 & 0.425 & 0.366 & 0.377 & 0.374\\
10 & 0.476 & 0.430 & 0.381 & 0.351 & 0.349 & 0.426 & 0.417 & 0.369 & 0.361 & 0.349\\
11 & 0.425 & 0.429 & 0.400 & 0.398 & 0.385 & 0.430 & 0.391 & 0.346 & 0.354 & 0.347\\
12 & 0.523 & 0.454 & 0.446 & \textbf{0.439} & \textbf{0.431}  & 0.469 & 0.433 & \textbf{0.440}  & 0.402 & 0.385 \\ [1ex] 
\hline 
\end{tabular}
}
\caption{20NG GMM (left) and IMDB GMM (right) clustering on BERT-base attention weights on the left and the right. The values depict coherence results. VB refers to the vanilla BERT-base model and FT refers to the fine-tuned version. The number followed by VB and FT refers to the number of clusters specified in the GMM model.} 
\label{20ng_bert_clustering}
\end{table*}

\begin{table*}[t]
\centering 
\scalebox{0.6}{
\begin{tabular}{c c c c c c c c c c c} 
\hline\hline 
Layer & VD30 & VD50 & VD100 & VD150 & VD200 & FT30 & FT50 & FT100 & FT150 & FT200 \\ [0.5ex] 
\hline 
1 &  0.504 &  0.497 &  0.518 &  0.515 &  0.521 &  0.502 &  0.508 &  0.511 &  0.517 &  0.513 \\
2 &  0.509 &  0.509 &  0.510 &  0.514 &  0.515 &  0.511 &  0.518 &  0.510 &  0.507 &  0.508 \\
3 &  0.514 &  0.514 &  0.508 &  0.508 &  0.510 &  0.507 &  0.503 &  0.503 &  0.499 &  0.507 \\
4 &  0.516 &  0.509 &  0.516 &  0.517 &  0.513 &  0.502 &  0.502 &  0.504 &  0.503 &  0.505 \\
5 &  0.548 &  0.550 &  0.551 &  0.544 &  0.549 &  0.544 &  0.550 &  0.546 &  0.543 &  0.544 \\
6 &  \textbf{0.593} &  \textbf{0.572} &  \textbf{0.572} &  \textbf{0.573} &  \textbf{0.568} &  \textbf{0.573} &  \textbf{0.576} &  \textbf{0.573} &  \textbf{0.571} &  \textbf{0.571} \\
\hline 
\end{tabular}
\begin{tabular}{c c c c c c c c c c c} 
\hline\hline 
Layer & VD2 & VD5 & VD10 & VD20 & VD30 & FT2 & FT5 & FT10 & FT20 & FT30\\ [0.5ex] 
\hline 
1 &  0.231 &  0.258 &  0.249 &  0.250 &  0.251 &  0.219 &  0.224 &  0.226 &  0.248 &  0.238 \\
2 &  0.166 &  0.211 &  0.212 &  0.224 &  0.230 &  0.255 &  0.225 &  0.231 &  0.232 &  0.238 \\
3 &  0.231 &  0.244 &  0.235 &  0.251 &  0.244 &  0.253 &  0.225 &  0.230 &  0.235 &  0.234 \\
4 &  0.151 &  0.228 &  0.204 &  0.228 &  0.234 &  0.170 &  0.223 &  0.218 &  0.219 &  0.217 \\
5 &  0.317 &  \textbf{0.347} &  0.307 &  0.270 &  0.264 &  0.252 &  0.268 &  \textbf{0.274} &  \textbf{0.272} &  0.261 \\
6 &  \textbf{0.334} &  0.327 &  \textbf{0.325} &  \textbf{0.317} &  \textbf{0.312} &  \textbf{0.288} &  \textbf{0.270} &  0.267 & 0.254 &  \textbf{0.264} \\[1ex] 
\hline 
\end{tabular}
}
\caption{20NG GMM (left) and IMDB GMM (right) clustering on DistilBERT attention weights. The values depict coherence results. VD refers to the vanilla DistilBERT model and FT refers to the fine-tuned version. The number followed by VD and FT refers to the number of clusters specified to the GMM model.} 
\label{20ng_distibert_clustering} 
\end{table*}

\section{Discussion}\label{discussion}
%
We have computed cluster coherence values on two different datasets. Given two clusters with their respective coherence values. If one cluster's coherence value is higher than the other, the one with the higher coherence values is regarded as a coherent cluster, for instance, in the case of text, the tokens in the coherent clusters tend to be semantically associated with each other. In both LDA and NMF models, we have varied the number of topics to demonstrate the impact of topic clusters. In Table~\ref{20ng_coherence} we present the topic coherence results in the 20 Newsgroups and IMDB datasets for the LDA and NMF models. We observe that in the LDA model when the number of topics is 20, we obtain the best coherence value of 0.518 in the 20NG dataset. In the case of the NMF model, the best coherence value is when the number of factors is 30 with 0.504 in the 20NG dataset. In the IMDB dataset, we also obtain the best coherence value when the number of topics is 20 in the LDA model with a value of 0.461 and the NMF model gives us the value 0.300 when the number of factors is 20.

\emph{PLM \& 20NG dataset:} In the case of the vanilla BERT-base model in Table~\ref{20ng_bert_clustering} (left), i.e., 20NG dataset, we notice that when the number of soft attention clusters is 50 there is some comparable performance with the coherence results. Precisely, we read from the table that for VB50 the coherence value is 0.503 in layer 11. This coherence value is numerically close to 0.518 when the number of topics is 20, and in the case of the NMF model, it is approximately equal to 0.504 when the number of factors is 30. This suggests that both LDA and vanilla BERT-base attention word clusters are semantically coherent when the number of soft clusters is 50. We also notice that the contextual layers are mainly playing a key role in modelling such semantically close words, i.e., layer 11. When we refer to the word overlaps in Table~\ref{bert_word_overlaps}, we notice that the top 20 word overlaps are also consistent with the BERT-base model in layers 7, 8, 9 and 11. It means that out of 20 words, there are 17 overlapping words.

Upon comparing the results of the fine-tuned version of the BERT-base model where the fine-tuning was done on the classification task, we notice that soft clusters 50 and 100 in Table~\ref{20ng_bert_clustering} lead to comparable coherence performances obtained by the LDA and NMF models in Table~\ref{20ng_coherence}. Precisely, we read from the table that when the number of clusters is 50 and 100, we obtain the coherence value of 0.508 and 0.503, respectively that again are numerically comparable to 0.518 in the coherence table for LDA and 0.504 for the NMF model, i.e., Table~\ref{20ng_coherence}. While it would be ideal to have these coherence results be equal, such results are difficult to obtain considering noise in the data and the randomness involved when initiating the training process of these semantic models. What is interesting in the case of the fine-tuned version of the BERT-base model is that two layers show comparable coherence performances and both these layers learn contextual information.

When we look at the topic associated with ``computing technology'' in the 20NG dataset, we noticed that words such as ``organisation'', ``com'', and ``nntp'' were among the overlapping words which suggest that both BERT and LDA learn thematically the same words. While it can be argued that even simple clustering algorithms such as k-means might generate clusters that are coherent and with high-overlapping words, we have found out that k-means does not lead to coherent clusters and the word overlap count was also very low, for instance, in most cases we found the word overlap values to be sometimes 1, and most often, 0.

In the case of the vanilla DistilBERT model presented in Table~\ref{20ng_distibert_clustering} (left), we notice that the higher layers demonstrate the highest soft cluster coherence results. What we notice is that the contextual layers show a higher degree of cluster coherence comparable to performance with the LDA model than with the NMF model in Table~\ref{20ng_coherence}, for instance, the vanilla DistilBERT version with 200 soft clusters shows a relatively comparable performance when compared with the LDA model in Table~\ref{20ng_coherence}. It can be argued that in terms of the absolute numbers the results in Table~\ref{20ng_distibert_clustering} are much higher than in Table~\ref{20ng_coherence} when we only look at the highest DistilBERT layers values. One of the reasons is that different pre-processing strategies have been chosen in both models. However, this was unavoidable because including stop words in the PTM models would result in noisy topics. Note that other layers such as Layer 4, soft cluster 30, in the case of the vanilla DistilBERT model compare well with the LDA coherence results. Layer 4 in the case of the DistilBERT model compares reliably with the soft cluster 30 when we consider the NMF model.

\emph{PLM \& IMDB dataset:} In the IMDB dataset, Table~\ref{20ng_coherence} presents the ideal coherence value when the number of topics/factors is 20 for the LDA and the NMF models. For the LDA model, the coherence value is 0.461 and for the NMF model, the coherence value is 0.300. Referring to Table~\ref{20ng_bert_clustering} (right), we see that the comparable LDA value is obtained in layer 6 in the vanilla BERT-base version when the number of soft clusters is 10. In the fine-tuned version, we see the comparable value in layer 8 when compared to the LDA model and when the number of soft clusters is 150. If we consider topic 30 in Table~\ref{20ng_coherence}, we notice two comparable values in Table~\ref{20ng_bert_clustering} in layer 12 which is a layer that captures contextual information more than any other layer when the vanilla soft clusters are 20 and 30. 

In Table~\ref{bert_word_overlaps}, most word overlaps occur in layers 5, 9, 11, and 12 and these results are consistent with the 20NG results where higher contextual layers have the maximum word overlap. We also notice that layers 6 and above have the most ideal coherence values indicating that if the clusters are coherent, they also have maximum word overlaps. It means that these clusters share common words. In DistilBERT, in Tables~\ref{20ng_distibert_clustering} and \ref{bert_word_overlaps} we see that the NMF model tends to show comparable coherence values in the higher layers. In Table~\ref{bert_word_overlaps}, we observe that the word overlaps are fairly uniformly distributed across layers. While the lower layers have shown to have maximum overlaps, we can notice that the upper layers too have a word similar overlaps. However, their coherence values are not comparable. It is because IMDB instances are short noisy sentences where the model seems to be performing not very reliably unlike the 20NG dataset. What is also noticeable from the results is that the fine-tuned version of the DistilBERT model does not show comparable coherence performance when compared with the NMF model. This could suggest that classification fine-tuning helps DistilBERT lose the latent topic information.

In summary: 1) the attention mechanism is an important component in the PLMs that help capture some patterns that are also captured by PTMs. 2) there is correspondence between the coherence results obtained from PLMs and PTMs because in most cases we obtain comparable coherence performance. 3) in PLMs, there are high word overlaps in the contextualised layers and clusters of words obtained from PTMs. 4) in most cases, it is the contextualised layer that captures the most commonality with PTMs.

One of the limitations of our work is that it does not experiment with other language models very different from BERT such as XLNet~\cite{yang2019xlnet} and GPT-3~\cite{floridi2020gpt} to ascertain that similar conclusions could be also derived from them. However, what is important to note is that our conclusions point toward the importance of the attention mechanism rather than the way pre-training is done or the size of the dataset that has been used to pre-train the model, or the model design. We also have to verify whether the results are generalizable to even larger models such as BERT-large which requires much more computational resources to conduct this study.

\begin{table}[t]
\centering
\begin{tabular}{c c c} 
\hline\hline 
Layer & 20NG & IMDB  \\ 
\hline 
1 & 16 & 14 \\ 
2 &  16 & 13 \\
3 &   16 & 12 \\
4 &   16 & 14 \\
5 &   16 & \textbf{17} \\
6 &   16 & 14 \\
7 &   \textbf{17} & 12 \\
8 &   \textbf{17} & 12 \\
9 &   \textbf{17} & \textbf{17} \\
10 &   16 & 11 \\
11 &   \textbf{17} & \textbf{17} \\
12 &  16 & \textbf{17} \\ [1ex] 
\hline 
\end{tabular}
\quad
\begin{tabular}{c c c} 
\hline\hline 
Layer & 20NG & IMDB  \\ 
\hline 
1 & \textbf{12} & \textbf{10}\\ 
2 &  \textbf{12} & \textbf{10} \\
3 &  9  & \textbf{10} \\
4 &  \textbf{12}  & \textbf{10} \\
5 &  \textbf{12}  & \textbf{10}\\
6 &  \textbf{12}  & 9 \\[1ex] 
\hline 
\end{tabular} %
\caption{BERT (left) and DistilBERT (right) attention word overlap with LDA.}
\label{bert_word_overlaps}
\end{table}

\begin{figure}[t]
    \centering
    \includegraphics[scale=0.5]{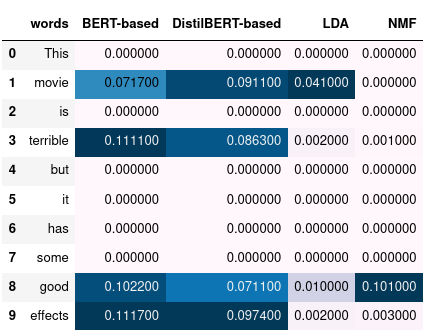}
    \caption{Illustrating attention using a sentence from the IMDB dataset as an example. We have presented these results from the BERT-base layer 11 and DistilBERT-based layer 5. The number of topics/factors in the case of PTM is 20. The figure is used to demonstrate that these models tend to focus on relevant tokens within their context and assign lower weights to general tokens such as stopwords.}
    \label{attention_illustration}
\end{figure}


We show another finding through Figure~\ref{attention_illustration} where we demonstrate the importance of the attention mechanism and how topic weights (probabilities) and attention weights tend to focus on the same words in a given context. To generate the figure, we have taken an example from the IMDB dataset. In the BERT-base model, layer 11 is examined because it is the contextual layer and has the highest word overlaps in Table~\ref{bert_word_overlaps}. In the case of the DistilBERT-base model, we have selected layer 5 given that it is one of the contextual layers and has one of the highest word overlaps in Table~\ref{bert_word_overlaps}. We have selected the number of topics as 20 and the number of NMF factors as 20 which is based on the results obtained in Table~\ref{20ng_coherence}. What we observe from the figure all the models tend to focus on the relevant keywords in the context, for instance, we observe that PLMs focus on the words such as ``good'', ``effects'', ``terrible'', ``movie'' that are relevant to the movie and the PTMs too tend to focus on the same tokens in this context. What we learn from the figure is that PTMs and PLMs, while they are different, both tend to focus on the relevant words in a given contextual window. This figure helps us to draw some relationships between the attention weights and the topic probabilities in that they focus on the important words only. We also notice that common words such as stopwords are given less weightage by the models.

While the authors in \cite{thompson2020topic} have found out that the word clusters obtained from some PLMs tend to cluster the contextualised word vectors that resemble what is learned by a topic model, our result suggests that it is the attention mechanism that is playing a key role in obtaining such results which is the key contribution of our work. It can also be argued that the contextualised token embeddings obtained from a PLM model can lead to almost similar conclusions, in this work, we wanted to explicitly study the role of the attention weights.

\section{Conclusions}\label{conclusions}
%
Topic modelling has remained a dominant modelling paradigm in the last decade with several topic models developed in the literature~\cite{zhao2021topic}. Topic models were not only modelled using Bayesian statistics but also linear algebra-based such as the NMF model. While both these models are formulated differently, they both tend to exhibit similar clustering properties. With the development of PLMs, these models have now taken over the landscape in text mining and NLP because they have outperformed existing baselines. Recent research points out that word-level clustering on BERT embeddings results in word clusters that share a close relationship with those discovered using topic models. As a result, this motivated us to study the reason which component in the language model helps capture such topic information when the model has not been explicitly designed to model latent word topics. Through probe tasks, we find that it is the attention mechanism that plays a key role in modelling word patterns that resemble something that is also discovered using topic models. We strongly believe that our work helps add further insight into the relationships between topic models and PLMs including the role that is played by the attention mechanism in the language model. In the future, we will conduct a thorough theoretical analysis to find out the key theoretical similarities between a topic model and a PLM. We will also study how different PLMs other than those that are based on BERT encode latent topics using attention weights.

Our results are not only applicable to NLP and document modelling fields in general, but the results are also relevant to information retrieval. For instance, in an information retrieval setting, we can only use features obtained from PLMs to retrieve relevant documents without having to worry about latent topics features that would potentially increase the number of features that might even degrade the performance of an information retrieval engine. Besides that, we may be injecting more redundant features into the information retrieval model. Topic models have been shown to improve information retrieval results and PLMs have been shown to demonstrate even better results. This could be because PLMs already have encoded a variety of features in their rich vector space that includes latent topics. As a result, the improvement that we see also comes from topics implicitly encoded in the PLM attention vectors. We thus believe that our paper will have a significant impact in the information retrieval field too.

\bibliographystyle{splncs04}
\bibliography{references}
\end{document}